\documentclass[10pt,letterpaper]{article}
\usepackage[top=0.85in,left=1in,right=1in,footskip=0.75in]{geometry}
\usepackage{amsmath,amssymb}

\usepackage{changepage}

\usepackage{textcomp,marvosym}

\usepackage[superscript]{cite}

\usepackage{nameref}
\usepackage[hidelinks]{hyperref}

\usepackage[nopatch=eqnum]{microtype}
\DisableLigatures[f]{encoding = *, family = * }

\usepackage[table]{xcolor}

\usepackage{array}

\newcolumntype{+}{!{\vrule width 2pt}}

\newlength\savedwidth

\definecolor{titleblue}{HTML}{273B81}

\raggedright
\usepackage[aboveskip=1pt,labelfont=bf,labelsep=period,justification=raggedright,singlelinecheck=off]{caption}
\renewcommand{\figurename}{Figure}

\makeatletter
\renewcommand{\@biblabel}[1]{\quad#1.}
\makeatother

\usepackage{lastpage,fancyhdr,graphicx}
\usepackage{epstopdf}
\fancyheadoffset[L]{2.25in}
\fancyfootoffset[L]{2.25in}
\begin{document}
\vspace*{0.2in}

\begin{flushleft}
\begin{center}
{\fontsize{22}{26}\selectfont\bfseries\color{titleblue} TrackStudio: An Integrated Toolkit for Markerless Tracking\par}

\vspace{0.35cm}

{\bfseries
Hristo Dimitrov\textsuperscript{1*},
Viktorija Pavalkyte\textsuperscript{1},
Giulia Dominijanni\textsuperscript{1,2},
Tamar R. Makin\textsuperscript{1}
\par}
\end{center}

\vspace{0.4cm}

{\bfseries
\textsuperscript{1}MRC Cognition and Brain Sciences Unit, University of Cambridge\\
\textsuperscript{2}Center for Neuroprosthetics and School of Engineering, École\\
Polytechnique Fédérale de Lausanne\\
*For correspondence: hristo.dimitrov@mrc-cbu.cam.ac.uk
}

\vspace{0.5cm}

\end{flushleft}


%
\section*{Abstract}
Markerless motion tracking has advanced rapidly in the past 10 years and currently offers powerful opportunities for behavioural, clinical, and biomechanical research. While several specialised toolkits provide high performance for specific tasks, using existing tools still requires substantial technical expertise. There remains a gap in accessible, integrated solutions that deliver sufficient tracking for non-experts across diverse settings.

TrackStudio was developed to address this gap by combining established open-source tools into a single, modular, GUI-based pipeline that works out of the box. It provides video recording preprocessing, recording synchronisation, automatic 2D and 3D pose estimation, and visualisation without requiring any programming skills. We supply a user guide with practical advice for video acquisition, camera calibration, video synchronisation, and experimental setup, alongside documentation of common pitfalls and how to avoid them.

To validate the toolkit, we tested its performance across three environments using either low-cost webcams or high-resolution cameras, including challenging conditions for body position, lighting, space, and obstructions. Across 76 participants, average inter-frame correlations exceeded 0.98 and average triangulation errors remained low ($<13.6mm$ for hand tracking), demonstrating stable and consistent tracking. We further show that the same pipeline can be extended beyond hand tracking to other body and face regions. TrackStudio provides a practical, accessible route into markerless tracking for researchers or laypeople who need reliable performance without specialist expertise.


%

\clearpage
\newgeometry{top=0.85in,left=1in,right=1in,footskip=0.75in}

\section*{Introduction}
Tracking human movements has been critical to the vast majority of research on motor control and neuromotor disorders. For example, motion capture has been pivotal in defining the fundamental biomechanics of human gait and movement coordination~\cite{Winter2009} as well as characterising fine motor impairments in Parkinson’s disease~\cite{Roemmich2013}. Nevertheless, motion tracking has often required specialised equipment and expertise to collect annotated movement data. Since the advent of optoelectronic systems that record reflective markers placed on body landmarks, optical tracking has become the gold standard for movement tracking due to its high precision~\cite{Scataglini2024,Kanko2021,vanderKruk2018}. However, the use of optical tracking requires specialised equipment and tends to be restricted to dedicated environments (research labs, production studios, etc.). Optical tracking entails further technical limitations, such as (relative) high cost and required technical skills for an effective setup. Furthermore, the use of markers represents a fundamental limitation as they require constant visibility and precise placement, both of which will affect the system's accuracy and increase setup complexity. Other alternatives, such as inertial measurement units (IMUs)~\cite{Filippeschi2017,Garca-de-Villa2023} or magnetic tracking~\cite{Franz2014} have been developed to address some of these issues e.g. reducing cost and requirements for marker visibility. However, these approaches have not gained broader adoption, as their sensitivity to environmental interference (metals and magnets), tendency for error drift, and need for specialised expertise continue to pose barriers.

\begin{figure}[!h]
\includegraphics[width=\textwidth]{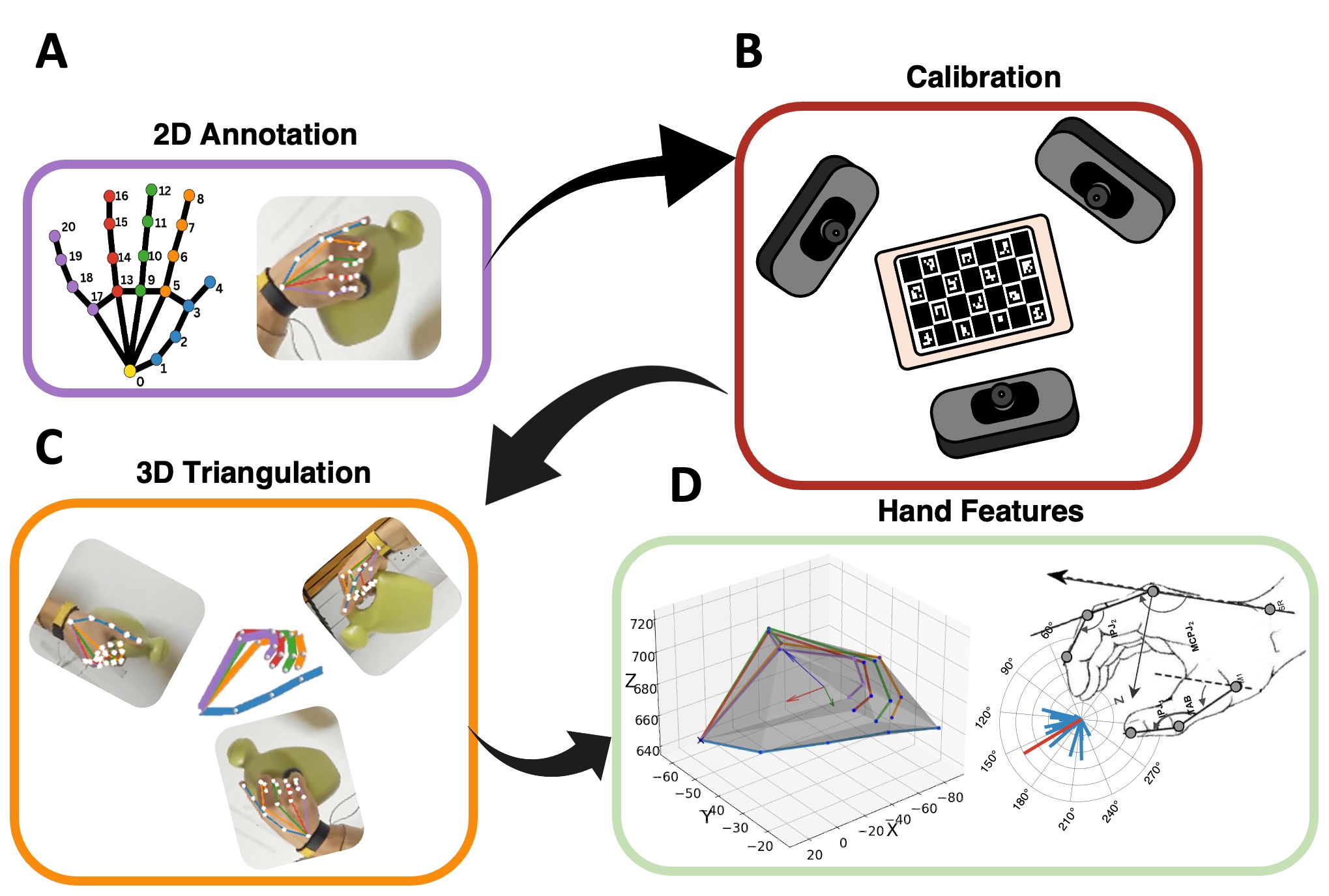}
\caption{A) 2D Hand Annotation based on Mediapipe Hand Landmarks. B) Example of a calibration process, required to obtain 3D information. C) Exemplary results of 3D data annotation. The displayed video frame has the hand landmarks labelled onto the original video. In the middle, there is a projection of the 3D acquired data of that frame. D) Extracting hand features such as angles and volumes.}
\label{fig1}
\end{figure}

Markerless tracking (MLT), which is the use of plain video footage to identify and track key landmarks on the human body, offers an alternative to marker-based tracking~\cite{Mathis2020,Kanko2021,Scataglini2024}. Current progress in computer vision and machine learning~\cite{Wang2021,Cao2021} allows for full automation of the process without the need for manual video annotation, simplifying the processing and expanding its usage to people outside the field of computer vision. The process of MLT (illustrated in Figure~\ref{fig1}) begins in two dimensions (2D) with an automated detection of specific body parts in individual images or videos (e.g. the position of a wrist, elbow, or fingertip). These landmarks are identified by computer vision models, which effectively act as “virtual markers” which are placed consistently across video frames in a process called 2D Annotation. Due to readily available large, annotated image datasets, and because the task is constrained to image coordinates, several pre-trained, off-the-shelf tools now exist and make automated landmark detection more accessible. For instance, MediaPipe Hand Landmarker~\cite{Zhang2020}, provides robust hand landmark detection across images, video, and live camera streams, while OpenPose~\cite{Cao2021} can detect body, face, and hand key points in real time, even in multi-person settings. These 2D tools eliminate the need for manual marking (such as in optical tracking) and make pose estimation available without the need for creating new models.

For many applications, however, 2D information is not enough. While 2D tracking can determine where a joint appears in an image or video, it cannot capture depth, orientation, or the spacespatial relationships between body parts. This data might be critical for analyses of movement kinematics, such as calculating joint angles, measuring stride length, or extracting other biomechanically meaningful features. Obtaining this information requires reconstructing three-dimensional (3D) poses from multiple camera views. Conceptually, this relies on 3D triangulation – the process of locating a point in space by combining its 2D positions from different camera viewpoints. For triangulation to correctly relate 2D pixel coordinates to 3D, the cameras themselves must be precisely calibrated – that is, their position, orientation, and lens distortions must be known. To obtain these parameters, typically, a reference checkerboard with known dimensions is recorded (illustrated in Figure~\ref{fig1}) and then processed by a camera-calibration algorithm. Additional challenges for triangulation include ensuring that the same landmark is consistently identified across views (multi-view correspondence) and reducing noise in the reconstructed trajectories. Toolkits such as Anipose~\cite{Karashchuk2021} and Pose3D~\cite{Sheshadri2020} offer tools for 3D triangulation and camera calibration as an open-source solution. Anipose also provides video overlays of resulting virtual markers and can integrate with another existing tracker software called DeepLabCut~\cite{Mathis2018}, to enable tracking of external objects and certain animals. 

All these tools (MediaPipe, OpenPose, Anipose, DeepLabCut) eliminate the manual annotation process, simplify camera calibration, and improve consistency and reproducibility. Additionally, recent advancements in MLT make the process simpler and more accurate each year. For example, the Radical Live service~\cite{RADiCALMotion2025} removes the need for high-end computers due to its cloud service. New algorithms, such as ATHENA~\cite{Mulla2025} use novel triangulation methods and mechanisms to prevent hand-switching errors (accidentally swapping which hand is being tracked), which bring performance closer to optical tracking.

Despite the barriers of specialised equipment and machine learning accuracy being lifted by various recent packages and toolkits~\cite{Uhlrich2023,TensorFlow2024,MMPoseContributors2020}, MLT is still not fully accessible for general use. The necessity for users to have coding knowledge, the ability to diagnose installation issues, and the ability to navigate through inconsistent standards, presents a technical barrier for wider MLT adoption. For comparison, Nexus (Vicon, Oxford, UK), a popular software for optical tracking, provides an intuitive graphical user interface (GUI) with ready-to-use features without the need for any coding or installation troubleshooting. Here we aim to close this gap by presenting TrackStudio – a toolkit that provides an accessible software tool for people wishing to employ MLT without requiring an extensive technical background. This community open-source resource also enables experts to expand the toolkit through integrating different methods for annotating 2D, 3D, or computing camera calibration for a variety of videos.

The TrackStudio toolkit is a collection of custom-written code and open-source libraries, aimed at simplifying and optimising MLT. It is designed to provide easy access and adequate performance in generalised settings, which could be easily utilised by non-experts. In addition to MLT, the TrackStudio toolkit includes modules for video trimming, and visualisation of preprocessing steps, as well as 2D/3D video labelling to support accessible quality inspection. All these features are integrated into a Python GUI, allowing users to access and customise the different tools and their parameters without having to interact with or write any code. The only inputs the TrackStudio requires are video recording(s) using a simple web camera(s), and a recording of a calibration board if 3D MLT is required (instructions provided in the Appendix). To facilitate evaluation and demonstration of the toolkit, example videos are available on the TrackStudio GitHub repository (https://github.com/dimitrov-hristo/TrackStudio/tree/master/examples/videos), allowing users to explore its full functionality.

\section*{Materials and methods}
\subsection*{Toolkit Overview}

The TrackStudio toolkit, including the utilities for video trimming, visualisation, configuration file editing, and tool integration was developed by the lead author (H. Dimitrov). The pose estimation models used for 2D/3D annotation, video labelling, and calibration were sourced from established open-source libraries.

2D video annotation is performed using Google’s MediaPipe framework~\cite{Zhang2020}. This tool was chosen due to its robust and lightweight implementation, as well as the stable performance in a variety of settings (lights, background, occlusions, etc.)~\cite{Bittner2023,TonyHii2023,DeCoster2023}. However, the standalone 3D model has limitations regarding depth estimation~\cite{DeCoster2023,Amprimo2022}. Therefore, Anipose’s approach~\cite{Karashchuk2021} is utilised for 3D triangulation, video labelling, and integration capabilities with DeepLabCut (used for animal models and inclusion of custom markers~\cite{Mathis2018}). By leveraging the strengths of these open-source libraries, we aim to extract optimal performance in a variety of conditions, while maintaining modularity for future expansion in the fast-moving MLT field.

One of the biggest challenges for using tools such as MediaPipe~\cite{Zhang2020}, OpenPose~\cite{Cao2021}, FreeMoCap~\cite{Queen2024} or Anipose~\cite{Karashchuk2021} is their maintenance and the continuously changing behaviour of the packages they rely on. This results in installation issues - e.g., incompatibility between package versions within and across different tools as well as issues due to operating systems and silicon. Thus, the approach we take is to package and upload the environments required for the pose estimation models and utilities, which prevents all of these issues. The disadvantage of this approach is that the user is limited to a particular version of these open-source tools. However, this approach guarantees functionality and maintains installation simplicity, especially since these tools are not updated frequently, but the packages that they use are, which can break the system’s utility.

Another important consideration in the MLT process is the acquisition of high-quality, synchronised recordings, as the MLT performance is ultimately dependent on the quality of the video data. Factors like occlusions (either self or external objects or barriers), light conditions, movement speed, lens and recording distortions, clothing, or calibration quality can significantly affect the performance of any markerless (and in many cases optical) tracking. At the same time, we aim to provide the easiest and least demanding setup, which can be used in diverse environments. Thus, we have created a brief manual (see Appendix) aimed at illustrating how to install TrackStudio, record quality videos, perform MLT, and address challenges pertaining to common pitfalls of MLT. In addition, we provide experimental validation of the toolkit across 3 different setups, each introducing its own MLT challenges, as well as test videos on TrackStudio’s page (https://github.com/dimitrov-hristo/TrackStudio/tree/master/examples/videos).

\subsection*{Experimental Validation}
\subsubsection*{Validation Environments}

To test the TrackStudio toolkit in realistic and challenging settings, we recorded 76 participants (recruited in the period 29/03/2022 – 30/05/2025) in 3 different setups (seated, supine, and mixed), interacting with various objects (see~Table~\ref{table1}). Ethical approval for the data collections was obtained from the Cambridge Psychology Research Ethics Committee for the data used seated and mixed protocols and from the University College London Research Ethics Committee (UCL REC) for the data used in the seated protocol. All participants provided written informed consent before taking part. These setups vary in number and type of cameras, tracked body parts, space, light conditions, background, body and object positions, and tasks. This is aimed to provide an evaluation of the performance of the toolkit in variable conditions and occlusion scenarios. The first two environments (seated and supine) involve recordings completed with simple web cameras (Logitech Brio, 1080p, 60Hz) and evaluating the capabilities of a low-cost setup. The third environment involves high-resolution RGB cameras (FLIR Blackfly S BFS-U3-23S3C-C). It aims to evaluate the capabilities of the toolkit for tracking multiple body parts across different body positions (sitting and standing), in a confined space, with multiple tasks, obstructions, and multi-day recordings.

\begin{figure}[!h]
\includegraphics[width=\textwidth]{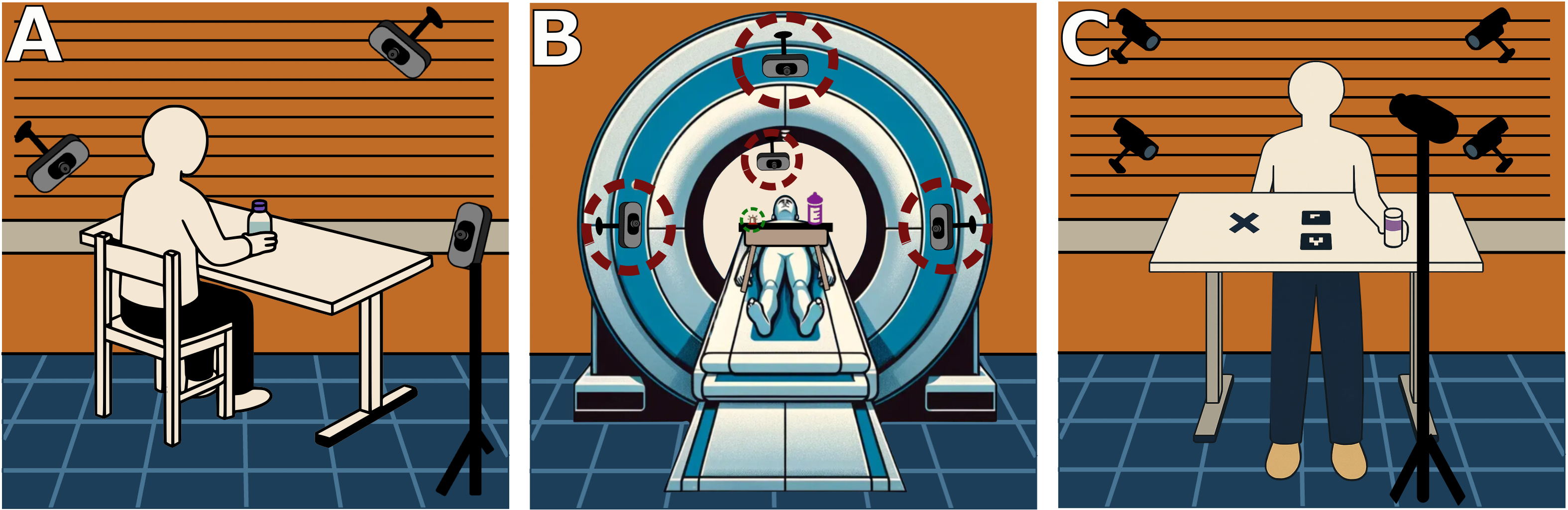}
\caption{Validation environments, illustrating the sitting setup with 3 web cameras (A), the lying down setup with 4 web cameras (B), and the mixed setup with 5 high-resolution cameras (C). For the lying down setup, web cameras are circled in red and LED light in green for clarity.}
\label{fig2}
\end{figure}

The seated environment tested the hand tracking performance across different tasks. The setup consisted of a table surrounded by three cameras, with participants sitting on a chair and performing object manipulation tasks (Figure~\ref{fig2}A). The data from this setup represents 45 people (27 female, mean age = 25.2 ± 4.47). Each participant performed a set of three tasks involving challenging manipulations of everyday objects~\cite{Dowdall2025}. Furthermore, participants also completed the same tasks wearing an augmentation device (The Third Thumb, Dani Clode Designs) on the tracked hand used for object manipulation. This hand-worn device introduces further occlusions to the MLT setup. Each task was repeated multiple times, yielding 10 minutes of video per task and a total of approximately 30 minutes of automatically labelled, tracked videos per participant and condition (with and without wearing an augmentation device). The three cameras were synchronised by the timestamps of the recording computer.

The supine environment tested hand tracking performance in a challenging body position in a constrained, low-light environment. The supine setup took place in a mock MRI room and consisted of four web cameras, with participants lying on the MRI bed being partially positioned inside the bore (only their head and neck). A height-adjustable table with an integrated LED light was mounted on the participants’ torso, serving as both a base for object placement and the designated start position of their hands. The LED light was used for camera synchronisation (Figure~\ref{fig2}C). Participants interacted with 10 different daily-life objects in two distinct ways – performing actions away from the body and towards the body. This setup produced data from 26 participants (14 female, mean age = 54.7 ± 12.8), comprising a total of 10 minutes of auto-labelled tracked video data per participant.

The mixed environment tested hand, arms and face tracking across multiple days and with position changes (sitting and standing) within a session. The setup used five high-resolution RGB machine-vision cameras distributed around the testing space. A red LED light was mounted in the upper corner of each camera to ensure video synchronisation. Participants performed seven different object manipulation tasks while sitting at a table and repeated a subset of three of these tasks while standing, with the table height adjusted accordingly (Figure~\ref{fig2}B). Here too, the hand augmentation device was worn for the majority (five) of the tasks during sitting and for all of the tasks during standing, imposing an additional occlusion challenge for tracking the biological fingers. This setup was tested on 5 participants (3 female, mean age = 25.7 ± 3.91). Each task was repeated multiple times, producing approximately 10 minutes of video per task. Hence, there was a total of ~70 minutes of automatically labelled, tracked video in the sitting condition and 30 minutes for the standing condition per participant per session. In total, there were 5 sessions in 5 consecutive days.


\newcolumntype{C}[1]{>{\centering\arraybackslash}p{#1}}

\begin{table}[!ht] 
\centering
\caption{\textbf{Summary of testing setups details.}}
\label{table1}
\begin{tabular}{|
C{0.1\textwidth}|
C{0.15\textwidth}|
C{0.1\textwidth}|
C{0.09\textwidth}|
C{0.13\textwidth}|
C{0.15\textwidth}|
C{0.09\textwidth}|
}
\hline
\rowcolor{gray!25}
\textbf{Testing Setup} & 
\textbf{Number of Participants} & 
\textbf{Number of Cameras} & 
\textbf{Number of Tasks} & 
\textbf{Recording across days} & 
\textbf{Tracking Data (per participant/session)} & 
\textbf{Tracking Type} \\
\hline

\cellcolor{gray!25}\textbf{Seated} &
45 (27 female, mean age = 25.2 $\pm$ 4.47) &
3 Webcams (Logitech Brio) &
3 &
Single session only &
30 min &
Hand \\
\hline

\cellcolor{gray!25}\textbf{Supine} &
26 (14 female, mean age = 54.7 $\pm$ 12.8) &
4 Webcams (Logitech Brio) &
2 &
Single session only &
10 min &
Hand \\
\hline

\cellcolor{gray!25}\textbf{Mixed} &
5 (3 female, mean age = 25.7 $\pm$ 3.91) &
5 High-Res Cameras (FLIR Blackfly S) &
7 (3 repeated while standing) &
5 &
70 min (sitting), 30 min (standing) per session &
Hand, arm, face \\
\hline
\end{tabular}
\end{table}

\subsubsection*{Preliminary workflow usability evaluation}
To assess the accessibility of the software for non-expert users, we conducted a brief structured usability evaluation with 10 participants reporting beginner-to-intermediate coding experience and no prior experience with the software. Before the session, participants were asked to download the software files and read the accompanying appendix manual. During the session, participants independently installed and configured the software using the example dataset, then completed the full processing workflow: automatic video trimming, camera calibration, 2D and 3D annotation, and 2D and 3D video labelling.

Task completion times were recorded throughout. To account for differences in computer speed, participant task time was calculated as total elapsed time minus processing time, allowing estimation of user-active time independent of hardware-dependent computation. 

\subsubsection*{Pipeline Performance Metrics}

To assess the performance of the pipeline, we have focused on quality metrics for MLT (for more details on ground truth comparison, see Tony Hii et al., 2023~\cite{TonyHii2023}). The quality metrics are: 1) inter-frame Pearson’s cross-correlation; 2) movement smoothness; 3) 3D error approximation. The inter-frame Pearson’s cross-correlation is at the level of the marker’s 3D vector magnitude and is performed between each 50ms of data, discarding the change of direction. We use 50ms intervals instead of consecutive frames and ignore correlations during changes of direction to avoid artificially high (minimal motion between frames) or low (natural movement reversal) values that do not reflect true tracking performance. The inter-frame cross-correlation per trial is computed by taking the median across all hand markers. Movement smoothness was quantified using the log dimensionless jerk (LDJ) metric. It represents the time- and amplitude-normalised integral of the squared jerk (the third derivative of position), providing a dimensionless measure of motion smoothness~\cite{Balasubramanian2012,Flash1985}. The LDJ per trial is computed by taking the median across all hand markers. As a guideline for interpreting LDJ, previous research indicates that simple actions/object manipulations result in LDJ in the range of 7–10~\cite{Bayle2023,Engdahl2019} and more complex movements/actions in the range of 11–14~\cite{Gulde2018}. Lastly, the 3D error approximation is representative of the amount of error incurred during the 3D triangulation. It is computed by calculating the error between the marker predicted by the 2D annotation and the 3D marker reprojected back into 2D space. To calculate this error in meaningful units (instead of camera pixels), we have converted pixels to mm by using the distance between the camera and the tracked body-part and the camera lens’ parameters. The error per trial is computed by taking the median across all hand markers. To contextualise acceptable magnitudes, state-of-the-art multiview triangulation methods typically achieve ~13–34 mm mean per-joint position error on human benchmarks~\cite{Bartol2022,Iskakov2019}. To capture further the number of large errors, we have calculated the percentage of errors that are above 10mm (for hand and face) and above 30mm (for elbows and shoulders), calculated per video frame and expressed as a percentage of the total number of frames. All of the MLT and metrics computation was performed on a standard laptop (Dell G5, Intel Core i5-8300H, Nvidia GeForce GTX 1060, 16GB RAM).

As TrackStudio combines Google’s MediaPipe for 2D annotations and Anipose for camera calibration and 3D reconstruction, prior research already exists on their accuracy compared to ground truth~\cite{Bittner2023,TonyHii2023,DeCoster2023,Amprimo2022,Hii2023,Amprimo2024,Sprague2025,Maggioni2025,Uribe2025,Karashchuk2021}. These studies provide a strong technical basis for the expected performance of the underlying methods. However, end-to-end tracking accuracy may still vary with camera configuration, calibration, movement characteristics, occlusion and environmental conditions. We therefore conducted a complementary target-referenced positional-accuracy assessment within the mixed environment recording configuration detailed above.

Participants performed a reaching task in which they sequentially touched eight fixed labelled targets (12 mm × 20 mm) with their fingertips. The targets were distributed across four pylons, with each pylon containing one target positioned at a higher and one at a lower height. For each expected target contact, we calculated the distance between the reconstructed fingertip position and the independently specified location of the relevant physical target.

A target contact was classified as successful when a 5-mm-radius sphere centred on the reconstructed fingertip intersected the relevant target region. We report the target-acquisition rate and, for unsuccessful contacts only, the median fingertip-sphere-to-target gap in millimetres. This benchmark provides an external, target-referenced assessment of positional accuracy in the task and recording environment used in the present study.

\subsubsection*{Statistical Analysis}

All statistical analyses were conducted in JASP (v0.18.3)~\cite{JASPTeam2025} using both frequentist and Bayesian approaches. For each setup, the three metrics (inter-frame cross-correlation, LDJ, and 3D error approximation) were reduced to a single median per subject per condition/task. Correlation coefficients were Fisher r-to-z transformed before inferential analyses. Descriptive statistics and figures are reported in the original correlation metric.

For the seated and supine setups, conditions were compared using paired t-tests (i.e., within-participant comparisons tested across participants) with accompanying Bayesian t-tests. For the Bayesian t-tests, the corresponding Bayes Factor ($BF_{10}$), defined as the relative support for the alternative hypothesis, was reported. We used the threshold of $BF_{10}<1/3$ as positive evidence in support of the null, consistent with previous research~\cite{Dienes2014,Wetzels2011}. The Cauchy prior width was set at the conventional default of 0.707~\cite{Dienes2014,Wetzels2011}. Furthermore, agreement between the two fixed conditions (augmentation device worn vs not worn or away vs towards) was assessed using the two-way mixed-effects, absolute-agreement, single-measure Interclass Correlation Coefficient (ICC(3,1)), with ICC $< 0.5$ = poor, 0.5–0.75 = moderate, 0.75–0.9 = good, and $> 0.9$ = excellent reliability.

For the mixed setup, for each body part (hand, face, arms), data were examined for consistency across days and between augmentation device conditions (device worn vs not worn) when sitting using both frequentist and Bayesian two-way repeated-measures ANOVAs (factors: Day × Device). If neither analysis indicated reliable effects or interactions, defined as non-significant frequentist results and $BF_{10} < 1$, data were considered stable and averaged across days and task types. This criterion was used solely to confirm measurement stability, as limited-sample Bayesian analyses often produce inconclusive Bayes Factors even when true differences are minimal~\cite{Dienes2014,vanDoorn2023,Keysers2020}. Descriptive statistics (mean ± SD) of the aggregated data are reported thereafter.

Statistical assumptions were assessed via Q–Q plots and Shapiro–Wilk tests applied to paired differences (for t-tests) or residuals (for ANOVA). Sphericity was evaluated using Mauchly's test, with Greenhouse-Geisser correction applied when violations were detected. Where distributional assumptions were materially violated, Wilcoxon signed-rank tests were applied, and z-statistic and Rank-Biserial Correlation were reported instead of t-score and Cohen’s d.

\section*{Results}

\begin{figure}[!h]
\includegraphics[width=\textwidth]{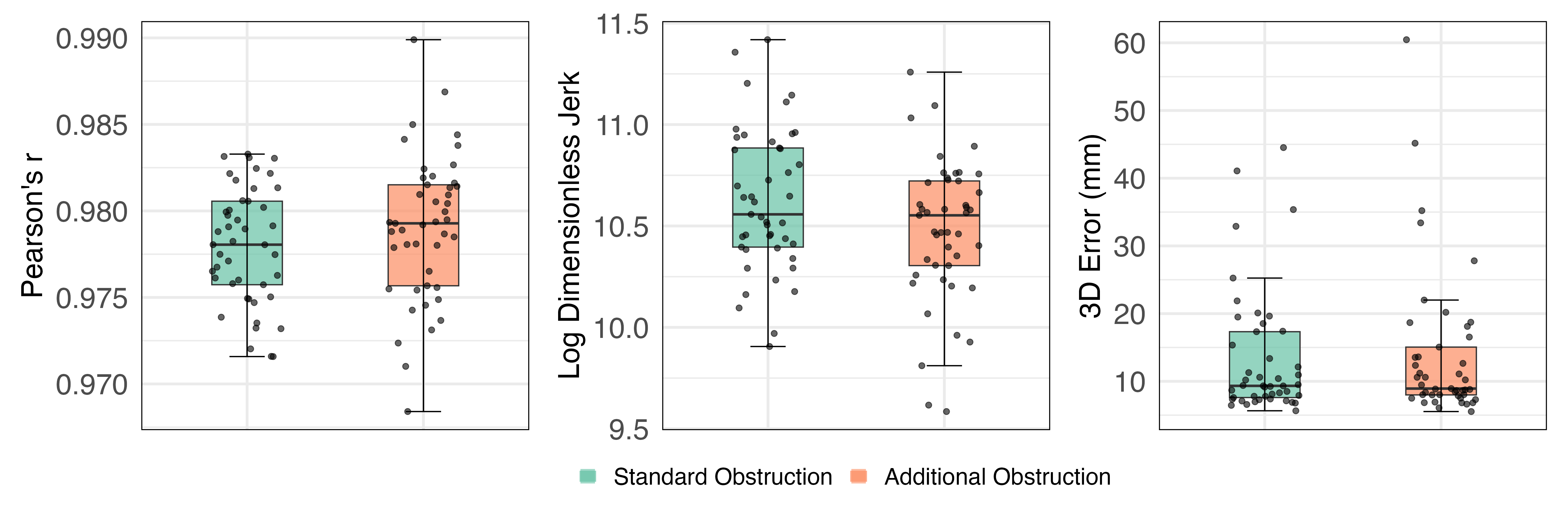}
\caption{Seated setup analysis plots, illustrating inter-frame cross-correlation, log dimensionless jerk, and 3D error, with colours signifying different conditions.}
\label{fig3}
\end{figure}

\begin{figure}[!h]
\includegraphics[width=\textwidth]{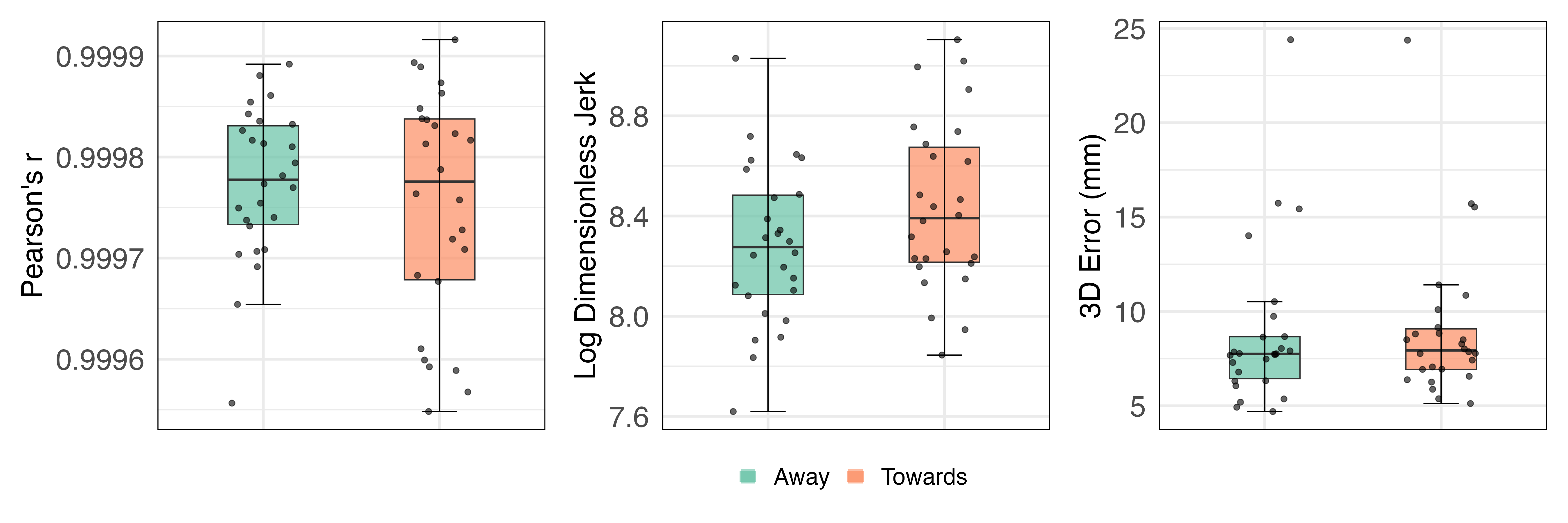}
\caption{Supine setup analysis plots, illustrating inter-frame cross-correlation, log dimensionless jerk, and 3D error, with colours signifying different conditions.}
\label{fig4}
\end{figure}

To assess TrackStudio’s performance with low-cost webcams under different conditions, we examined inter-frame cross-correlation, movement smoothness (LDJ), and re-projection error (absolute and percentage $>10$ mm) in the seated (Figure~\ref{fig3}) and supine (Figure~\ref{fig4}) setups. Cross-correlations were extremely high in both configurations (seated: r = 0.98 ± 0.004; supine: r = 0.999 ± 0.0001), indicating smooth transitions and minimal marker jumps, e.g. due to occlusions. LDJ values followed previously reported trends from other studies~\cite{Bayle2023,Engdahl2019,Gulde2018}: simple movements yielded lower LDJs (7–10)~\cite{Bayle2023,Engdahl2019} and more complex actions produced higher values (11–14)~\cite{Gulde2018}. The supine setup, involving a single object manipulation movement (low complexity), showed an average LDJ of 8.36 ± 0.33, while the seated setup, which required continuous performance of daily manipulation tasks (higher complexity), registered 10.55 ± 0.36.

The 3D error (incurred during triangulation) remained small across participants (seated: 13.6 mm ± 10 mm; supine: 8.95 mm ± 4.1 mm), with very few large errors (seated: 0.58\% ± 0.17\%; supine: 0.55\% ± 0.16\%). These errors are consistent with state-of-the-art results on human benchmarks~\cite{Bartol2022,Iskakov2019}. Furthermore, there was no significant effect on error from either the added obstruction in the seated setup (p = 0.37, z = 0.91, Rank-Biserial = 0.156, $BF_{10}$ = 0.23, ICC = 0.67) or the task type in the supine setup (p = 0.094, z = –1.689, Rank-Biserial = –0.379, $BF_{10}$ = 0.636, ICC = 0.95). Likewise, inter-frame cross-correlations did not differ between the two seated conditions (p = 0.08, t = –1.783, d = –0.266, $BF_{10}$ = 0.69, ICC = 0.44) or the two supine tasks (p = 0.248, t = –1.183, d = –0.232, $BF_{10}$ = 0.388, ICC = 0.56). As expected, LDJ differed significantly within both setups, reflecting the differing movement complexities (seated: p = 0.006, t = 2.886, d = 0.43, $BF_{10}$ = 6.01, ICC = 0.49; supine: $p < 0.001$, t = –4.151, d = –0.814, $BF_{10}$ = 90.3, ICC = 0.75).

\begin{figure}[!h]
\includegraphics[width=\textwidth]{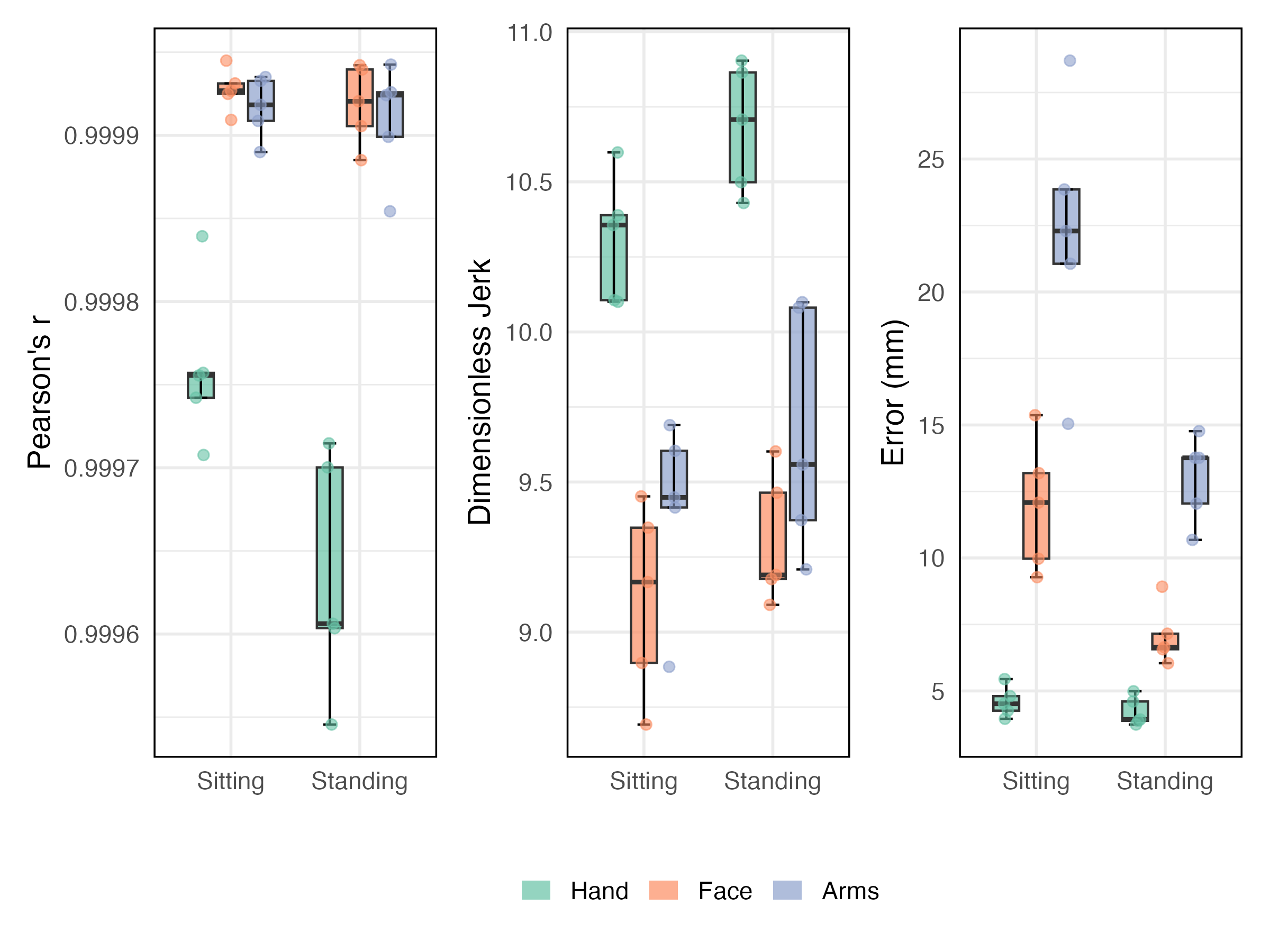}
\caption{Mixed setup analysis plots, illustrating inter-frame cross-correlation, log dimensionless jerk, and 3D error, with colours signifying different body parts and x-axis denoting different conditions.}
\label{fig5}
\end{figure}

To explore the benefits of higher-resolution cameras, we evaluated a mixed setup involving sitting and standing tasks, tracking the hand, face, and arms in a confined space over five consecutive days. Given the limited number of participants, these analyses were considered exploratory. No significant day-to-day or device-wearing differences were detected ($0.93 > p > 0.15$, $0.33< $ $BF_{10}$ $< 0.848$), suggesting no strong evidence for systematic changes across the recorded sessions. Therefore, Figure~\ref{fig5} shows participant-wise averages across days. Cross-correlations were again extremely high across all segments (0.999 ± 0.0001). Hand movements showed LDJ values (9.75 ± 0.62) similar to those in the seated webcam setup, reflecting comparable task structure. The 3D reprojection error for the hand was the lowest among all setups (4.41 mm ± 0.53 mm), with almost no large errors (0.0048\% ± 0.0016\%). Errors for face (9.52 mm ± 3 mm) and arms (17.6 mm ± 5.65 mm) were higher, likely reflecting the increased difficulty of tracking facial features and the larger anatomical variability of elbow and shoulder joint centres. Nevertheless, all reported errors remain within the range of values reported by current state-of-the-art approaches evaluated on human benchmarks~\cite{Bartol2022,Iskakov2019}.

In the mixed-camera setup, we further assessed target-referenced positional accuracy during the separate target-reaching task. Across 730 reaches across the 5 participants and 5 days, the 5-mm-radius sphere centred on the index fingertip intersected the relevant target region in 91\% of observations. For the remaining observations, the median gap between the fingertip sphere and the target boundary was 10.3 mm with interquartile range of 14.7mm.

Finally, all participants who took part in our preliminary assessment of the workflow usability completed successfully all the tasks from the first try. The average user-active completion time from installation to 3D video labelling was 04:44 minutes (range: 02:09 minutes-09:33 minutes), excluding computer processing time.

\section*{Discussion}

To move MLT beyond expert use, we aimed to shift the focus from demonstrations of technical accuracy alone towards expectations of usability, repeatability, and adaptability. The TrackStudio toolkit presented here demonstrates that a streamlined, pre-configured workflow can deliver stable performance across different tasks and environments without the need for technical expertise. It is not intended to improve or replace established pose-estimation or triangulation algorithms such as MediaPipe, Anipose, or DeepLabCut. Rather, its contribution is practical and workflow-based: it integrates these tools within an accessible graphical interface, with a simplified installation process and constrained analysis workflow that guides users from video preparation to pose estimation and visualisation. This is important because, in practice, successful MLT use depends not only on algorithmic performance, but also on the user’s ability to assemble compatible tools, manage software environments, define configuration files, prepare input videos appropriately, and transfer data between different stages of the workflow. By reducing this dependency on technical expertise, TrackStudio enables non-technical users to obtain reproducible outputs that are comparable to those achievable through more technically demanding implementations.

This positioning is important when comparing TrackStudio with other commonly used markerless tracking workflows. Tools such as ATHENA~\cite{Mulla2025}, DeepLabCut~\cite{Mathis2018}, Bonsai~\cite{Gonzalez2025}, OpenCap~\cite{Uhlrich2023}, and FreeMoCap~\cite{Queen2024} have substantially expanded access to markerless tracking, each with clear strengths. DeepLabCut/Anipose and Bonsai-based workflows offer high flexibility and are particularly valuable for custom key points, multi-camera reconstruction, or real-time experimental control, but they still require technical skills and choices around installation, annotation, configuration, and/or pipeline assembly. ATHENA provides an important step toward highly accurate no-annotation, GUI-based hand tracking, but it is currently focused on hand kinematics and requires short command-line inputs for installation and launching. OpenCap and FreeMoCap similarly reduce barriers to full-body motion capture yet are primarily organised around their own capture and processing workflows and require technical skills for installation and usage. TrackStudio fills a different gap: it provides a pre-configured, GUI-based route from video preparation to tracking, and visualisation, while remaining modular enough to incorporate outputs from tools such as DeepLabCut when specialist tracking is required.

Further strength of our pipeline lies not in outperforming task-specific implementations under ideal conditions, but in delivering dependable results across imperfect ones. Although direct comparison is limited, our target-referenced analysis within the mixed-camera setup, demonstrated broadly comparable results with prior markerless hand-tracking studies (reporting fingertip errors of approximately 11–21 mm)~\cite{Abdlkarim2024,Reimer2023}. However, in many research contexts, the practical barrier is not whether a model can achieve a millimetre precision, but whether the pipeline can be deployed with limited setup time and budget, non-specialist staff, and variable recording environments. The current findings suggest that TrackStudio occupies this middle ground effectively: it enables reproducible tracking without requiring control over hardware, room layout, or extensive expertise.

More broadly, the work highlights the importance of reducing not only algorithmic complexity but also procedural friction. Many existing pipelines fail to translate beyond expert laboratories because the barrier lies in configuration, installation, or uncertainty over how to align components. By packaging a predefined environment with integrated documentation, informed default settings, and processes for recovery from common pitfalls, TrackStudio offers practical accessibility to support a broader user base, including those who may face challenges when working with command line interfaces.

At the same time, the toolkit does not preclude expert use and upgrades to its utility. Its modular organisation allows substitution of models for 2D/3D tracking, calibration methods, and extension to additional body regions or multi-person tracking. The preliminary tests of face and arm tracking illustrate this adaptability, even if performance varies with anatomical complexity and marker definition. These differences are expected and highlight where task-specific refinement may be needed if sub-centimetre precision is critical.

There are, of course, boundaries to what a general-purpose workflow can resolve. The present approach and validation were limited to single-person recordings in relatively structured settings. Out-of-scope scenarios, including multi-person contexts, close physical interaction, or crowded scenes, introduce additional challenges that go beyond the single-person tracking problem evaluated here. In these settings, tracking errors may arise from identity switching across frames, incorrect assignment of landmarks to individuals, inter-person occlusion, or ambiguity when overlapping body parts are detected. As TrackStudio was designed and validated for single-person recordings, its applicability to multi-person contexts should therefore be tested separately and may require additional identity-tracking, camera coverage, or manual correction procedures. In addition, highly dynamic full-body movements, severe or unpredictable occlusions, or applications requiring clinical-grade kinematic accuracy may still benefit from tailored optimisation, additional cameras, or marker-based systems.

The current evaluation also highlighted practical limits related to recording space and hardware. All environments involved a single individual within an appropriately sized capture volume. When the recording space was not appropriately sized, as in the mixed setup, higher-end cameras were required to improve the field of view and shutter speed. Increasing the number of cameras may further improve coverage, but also increases demands on video capture, data transfer, synchronisation, and visibility of synchronisation cues (hence we used an LED per camera in the mixed setup). Nonetheless, the aim of the present work was not to eliminate the need for specialised solutions, but to make MLT realistically accessible in scenarios where it is currently under-used.

In the near future, we plan to expand the TrackStudio toolkit to different operating systems (Linux-based and MacOS-based), add advanced feature extraction tools, expand the methods for video recording synchronisation, and include multi-person tracking approaches. With this, we hope to expand further the utility and usability of the toolkit as well as improve the camera synchronisation process.

\section*{Conclusion}

By offering a pre-configured, GUI-based system, that supports diverse recording conditions without bespoke engineering, TrackStudio lowers the barrier to MLT adoption and facilitates experimentation rather than discouragement at the setup stage. In doing so, it also opens possibilities for educational activities, pilot work, and exploratory recording where the cost of failure is typically too high to justify time investment. TrackStudio does not represent an optimal system, but rather a practical one: a toolkit that lowers the barrier to entry while remaining adaptable.

\section*{Acknowledgments}

This work was supported by intramural funding from the Medical Research Council UK (MC\_UU\_00030/10) to TRM. We would like to thank Lucy Dowdall, Payton Kang, and Ema Jugovič for providing additional data for validation and Jonathan A. Michaels for contributing with guidance and ideas for setup improvements.

%
%

\bibliography{references_TrackStudio}

\clearpage
\appendix

\setcounter{figure}{0}
\renewcommand{\figurename}{Supplementary Figure}
\renewcommand{\thefigure}{\arabic{figure}}

\section*{Appendix – TrackStudio Manual}

\subsection*{Prerequisites}

For installation purposes, the user should already have Anaconda  software installed (https://www.anaconda.com/docs/getting-started/anaconda/install). 
Before installing the TrackStudio toolkit, download the toolkit’s files from GitHub (https://github.com/dimitrov-hristo/TrackStudio) and virtual environments files from FigShare (https://doi.org/10.6084/m9.figshare.30556418.v1). The unzipped toolkit files and the 2 virtual environment .tar files must all be in the same folder (see Supplementary Figure~\ref{suppfig1}).

\begin{figure}[!h]
\includegraphics[width=\textwidth]{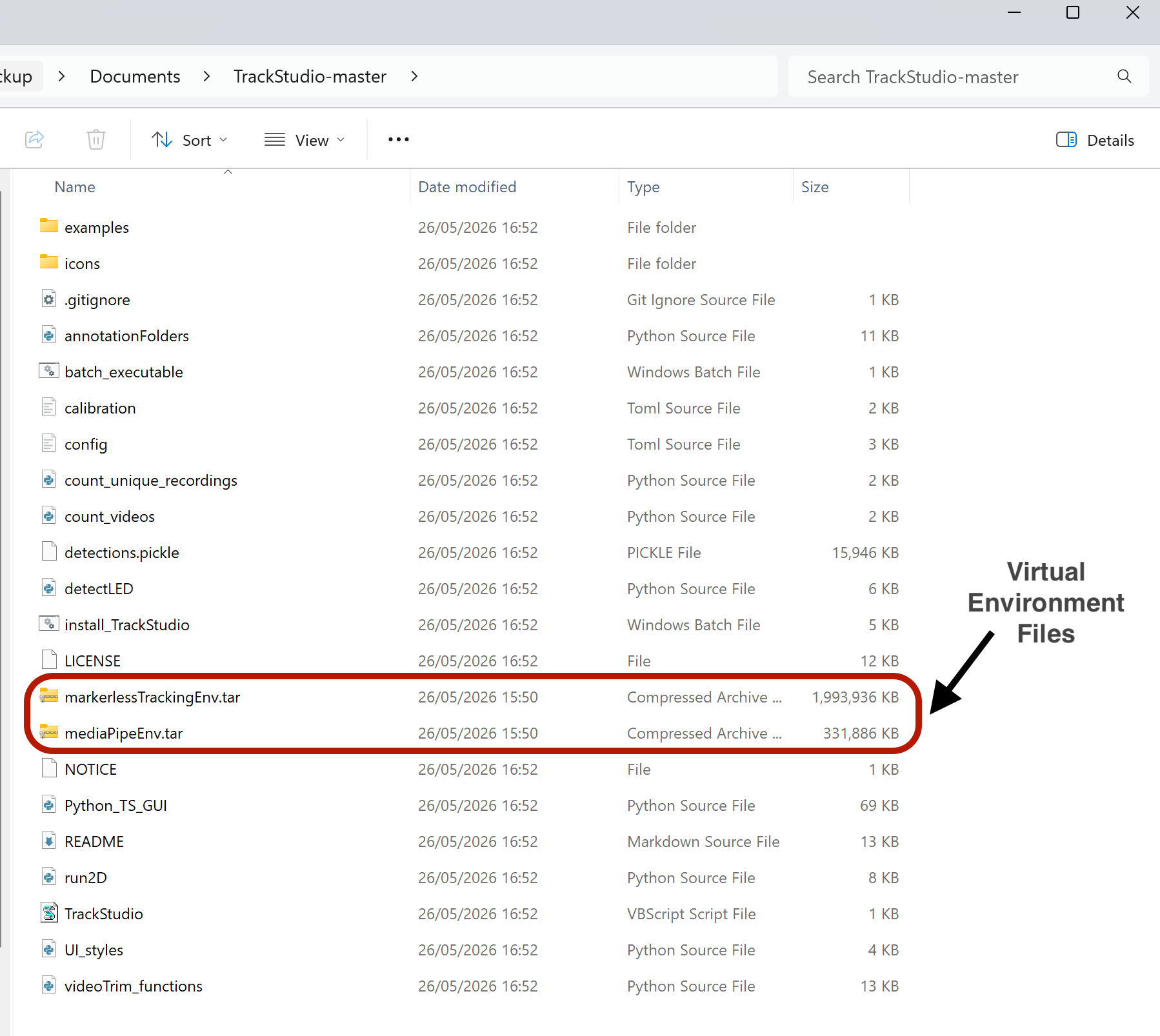}
\caption{TrackStudio master folder before installation.}
\label{suppfig1}
\end{figure}

\subsection*{Installation}

\begin{figure}[!h]
\includegraphics[width=\textwidth]{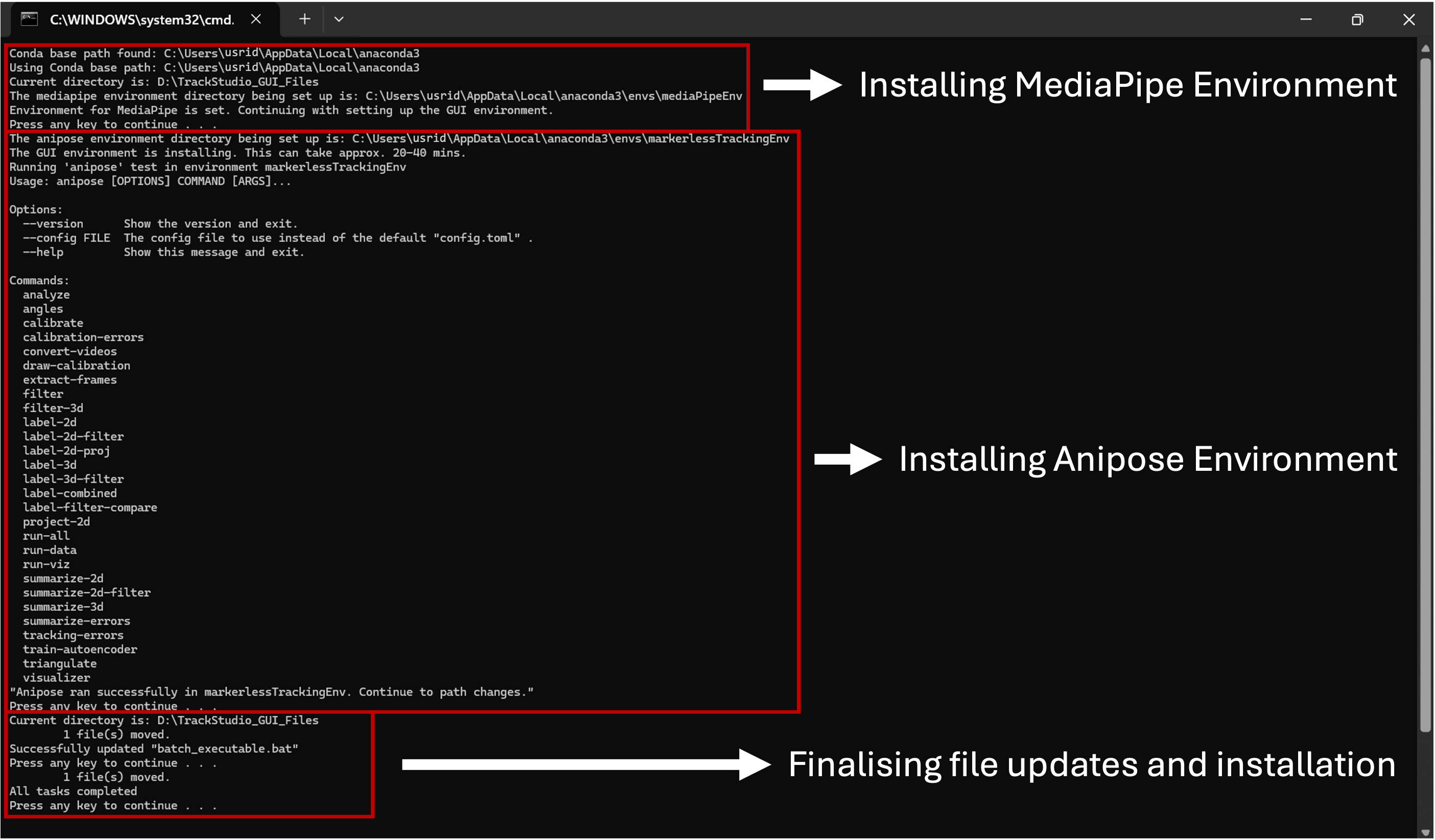}
\caption{Installation window visualisation, indicating where each step takes place.}
\label{suppfig2}
\end{figure}

To install TrackStudio you need to double-click the installation file (install\_TrackStudio.bat). The installation process may take 10-40 minutes. In the unlikely event the installer cannot automatically find your Anaconda folder, you will be prompted to enter the path. 
\underline{During installation (Supplementary Figure~\ref{suppfig2}), the installer will:}
\begin{enumerate}
\item Set up MediaPipe virtual environment → prompts “press any key” to continue.
\item Set up Anipose virtual environment (this is the longest step) and run a quick installation test → prompts “press any key.”
\item Update TrackStudio files → prompts “press any key.”
\item Finalise the setup → final “press any key,” and you are done.
\item Open the TrackStudio GUI by double-clicking the 'TrackStudio.vbs' file.
\end{enumerate}
Note: The first launch of the GUI (after a double-click of the 'TrackStudio.vbs' file) may take extra time while initial variables are created – wait until the GUI has been fully initiated before trying anything else. If the GUI does not start after 20-30s, try opening the Anaconda software as well as opening the 'batch\_executable.bat' and the 'TrackStudio.vbs' files as text files (e.g. in Notepad) to check if the paths within the 'batch\_executable.bat' and the 'TrackStudio.vbs' files have been updated correctly.

These paths should have been updated automatically. However, in case of an issue with the process:

\begin{itemize}
\item The path in the second line of 'TrackStudio.vbs' should have been updated from the default \verb|D:\Markerless_Tracking_GUI_git\batch_executable.bat| to the path of the folder where the track studio GitHub repository has been unzipped.
\item Similarly, the path in the last line of the 'batch\_executable.bat' should point towards the location of the 'Python\_TS\_GUI.py' file in your computer (where you've unzipped the GitHub repository) rather than the default \verb|D:\Markerless_Tracking_GUI_git\Python_TS_GUI.py| and the path to the batch file for activating Anaconda in the third line of the 'batch\_executable.bat' should point towards the one within your Anaconda installation.
\end{itemize}

After installation, the TrackStudio GUI can be launched by double-clicking the application file (TrackStudio.vbs). Advanced code information is provided on the GitHub page (https://github.com/dimitrov-hristo/TrackStudio).

There is an ‘examples’ folder provided for people to go through examples using the instructions from the ‘Usage’ section below and familiarise themselves with the GUI. The examples are located in \verb|TrackStudio/examples/videos| within the GitHub page. They contain \textbf{.mp4} recordings from 3 web cameras with file naming suffix ‘\textbf{-cam([A-Z])}’ of a person performing an action with their \textbf{right hand}. People can try the trimming functionality of the GUI by trimming videos by an LED light approach using the ‘raw-videos’ folder, or performing 2D annotation using the ‘trimmed-videos’ folder and subsequently performing camera calibration and 3D annotation using the ‘calibration’ folder and the subsequently generated files.

\subsection*{Before You Start: Recording \& Organising Video Recordings}

\subsubsection*{Video Recordings}

An easy way to record several cameras on one computer is to use OBS (Open Broadcaster Software, Lain Bailey, Michigan, US)~\cite{Bailey2017}. Open one OBS window per camera (e.g. three cameras → three OBS windows). Example OBS profiles and settings are available on the project’s GitHub.

\subsubsection*{Naming \& Saving Camera Recordings}

For a seamless MLT process, always save your video footage in a dedicated directory for video recordings. TrackStudio will mirror your folders (e.g. participant → task → trial), but the lowest-level folder (e.g. trial1) must contain video files only—no notes, images, or other files.
When recording cameras, give each camera a unique name using this format:

\begin{itemize}
\item Start with '\_' or '-'
\item Then 'cam'
\item Then a lowercase/capital letter (a-z)/(A-Z) or number (1-9)
\end{itemize}

3-camera examples: '-camA, -camB, -camC' or '\_cam1, \_cam2, \_cam3'.
You can set the suffixes used for each OBS window in Advanced Settings → Recording and File Formatting (see Supplementary Figure~\ref{suppfig3}).

\begin{figure}[!h]
\includegraphics[width=\textwidth]{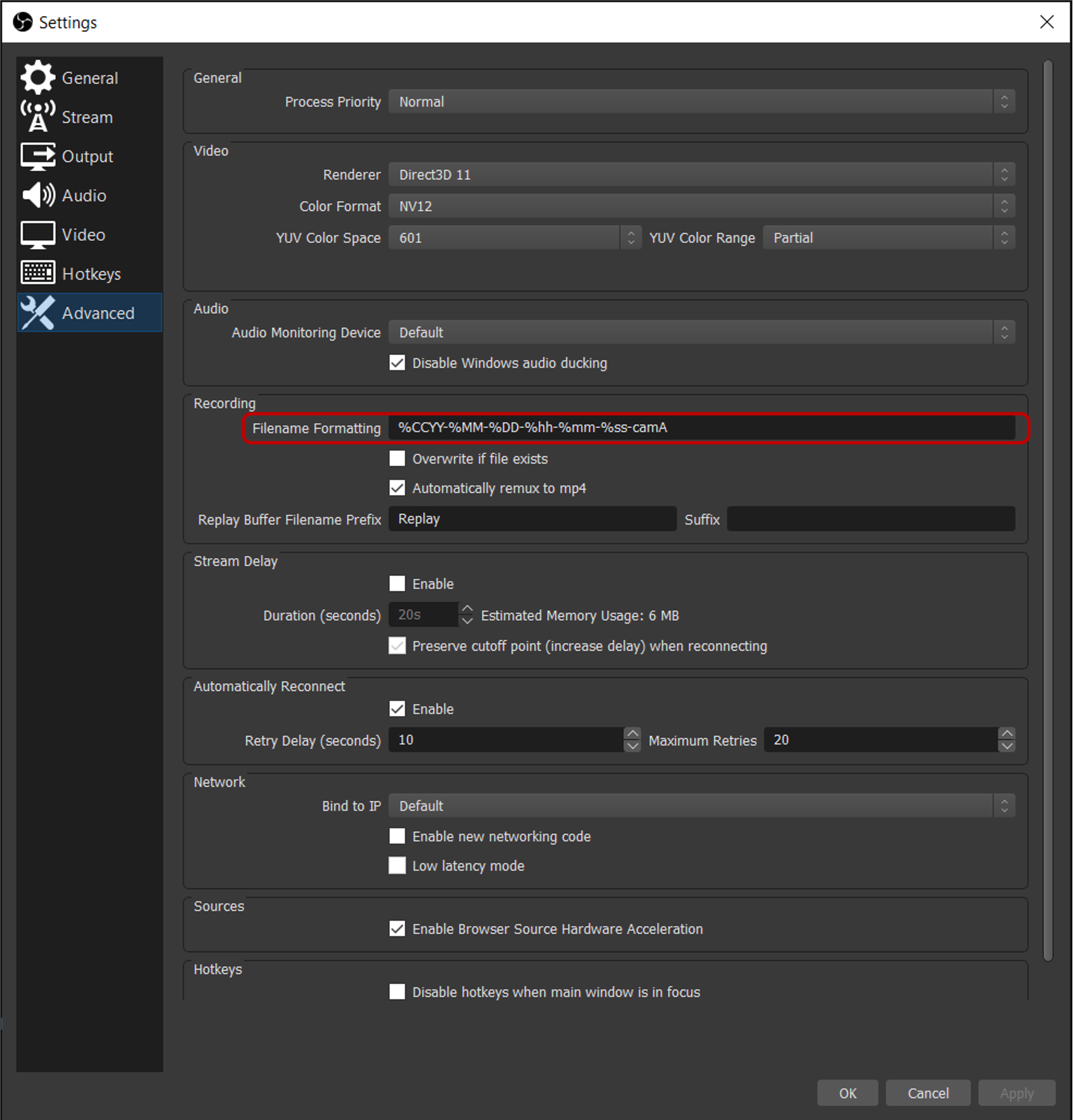}
\caption{Setting section of Open Broadcasting Software (OBS). Each camera recording should have a unique suffix at the end of the recorded file name.}
\label{suppfig3}
\end{figure}

\vspace{0.5em}

\par\noindent\underline{Optional: Using more than 3 webcams}

Although most users will not need this, if using more than three cameras or high-resolution/high-frame-rate cameras, please check:

\begin{itemize}
\item Cables/ports: Use cables that support your camera needs (e.g. high-resolution cameras may require USB-C 3.2 Gen 2 or better). Slow cables can cause dropped frames.
\item USB port limits: A laptop/desktop might have four USB-C ports but only two internal USB controllers ('boards'), meaning the connections are shared. If video lengths are not matching due to USB speed limitations, add a dedicated USB expansion card (desktop) or reduce the resolution/frame rate per camera.
\item Computer performance: Your computer must be able to process and display multiple live video streams and save them to its hard drive. If it cannot, video frames may skip at random (these cannot be recovered). \textit{Tips}: Record to a fast SSD, close other apps, reduce previews, and consider lowering FPS/resolution.
\end{itemize}

\subsubsection*{Synchronisation}

There are two different kinds of synchronisation:

\begin{enumerate}
    \item Event synchronisation – aligning videos with your task events (e.g. trial start/end).
    \item Camera synchronisation – making sure all cameras show the same moment on the same video frame.
\end{enumerate}

For 3D MLT, camera synchronisation is \textbf{essential}. Even a small delay (tens of milliseconds) can break the 3D estimation and corrupt the obtained results. For example, camera A might display an opened hand, while camera B already shows the hand closing. If you build a 3D view from those two video frames, the result will be corrupted as the two views are from different moments in time.

\vspace{0.5em}

\par\noindent\underline{Important note about global computer timestamps}

Global computer timestamps are useful for event synchronisation only (matching videos to your task). They do not make cameras aligned with each other. Each camera video is independent when it transfers the data to the computer. Hence, a simple start/stop timestamps can leave cameras out of synchronisation. Some softwares can extract timestamps for each camera separately, for example, LabRecorder (https://github.com/labstreaminglayer/App-LabRecorder)~\cite{Kothe2025} (see below).

\vspace{0.5em}

\par\noindent\underline{Recommended ways to synchronise cameras}

\begin{enumerate}
    \item Visual or audio cue (\textbf{recommended}):
Use an LED light that turns ON/OFF at known times (e.g. task start/transition/end) and is visible in every camera or use a sharp sound (a beep) recorded by all cameras. This provides a clear, shared timing marker you can see or hear later and verify it. If using sound, ensure audio is recorded in the video files.
    \item Software timestamping: 
Use of a software tool, such as LabRecorder~\cite{Kothe2025}, to extract separate timestamps per camera and event. Often adequate for slower motions or tolerant tasks, but can be inaccurate for time-critical applications, and cannot be verified.
\end{enumerate}

\vspace{0.5em}

\par\noindent\underline{Best practices for LED light synchronisation}

\begin{enumerate}
    \item Make sure every camera can see the LED clearly at all times. If not, add more LEDs that switch ON/OFF together. \textit{Tips}: for multiple LEDs, you can place an LED in the corner of each camera, which does not block the main view but is always visible.
    \item Use a dark background behind the LED (e.g. black cloth) that the ON/OFF changes are obvious.
    \item Do not let anything block the LED during ON/OFF events. If it is hidden in any view, synchronising and auto-trimming may fail. If a body part occludes the LED light, the video trimming tool may incorrectly interpret the skin tone as a change in LED colour (particularly when using a red LED). This may lead to the misidentification of ON/OFF events and result in inaccurate video trimming
    \item Simple LED wiring and an Arduino script example are available on GitHub (https://github.com/dimitrov-hristo/TrackStudio).
\end{enumerate}

\subsubsection*{Quick start checklist}

\noindent
\makebox[1.8em][l]{\fbox{\rule{0pt}{0.6em}\hspace{0.6em}}}%
\begin{minipage}[t]{\dimexpr\linewidth-1.8em\relax}
One OBS window per camera; you can use the provided OBS settings.
\end{minipage}

\vspace{0.5em}

\noindent
\makebox[1.8em][l]{\fbox{\rule{0pt}{0.6em}\hspace{0.6em}}}%
\begin{minipage}[t]{\dimexpr\linewidth-1.8em\relax}
Name cameras \texttt{-camA}, \texttt{-camB}, \texttt{-camC} 
or \texttt{\_cam1}, \texttt{\_cam2}, \ldots
\end{minipage}

\vspace{0.5em}

\noindent
\makebox[1.8em][l]{\fbox{\rule{0pt}{0.6em}\hspace{0.6em}}}%
\begin{minipage}[t]{\dimexpr\linewidth-1.8em\relax}
If using 4+ cameras: check cable speed, USB bandwidth/limits, and computer performance; a fast SSD helps.
\end{minipage}

\vspace{0.5em}

\noindent
\makebox[1.8em][l]{\fbox{\rule{0pt}{0.6em}\hspace{0.6em}}}%
\begin{minipage}[t]{\dimexpr\linewidth-1.8em\relax}
Choose a synchronisation method if required:
\begin{itemize}
    \item For 2D task alignment only $\rightarrow$ global computer timestamps; this does not align cameras.
    \item Basic camera synchronisation $\rightarrow$ software timestamping.
    \item Preferred camera sync $\rightarrow$ LED or sound that is visible/audible to every camera.
\end{itemize}
\end{minipage}

\vspace{0.5em}

\noindent
\makebox[1.8em][l]{\fbox{\rule{0pt}{0.6em}\hspace{0.6em}}}%
\begin{minipage}[t]{\dimexpr\linewidth-1.8em\relax}
Record a 10--15 s test. Check for no dropped frames and verify the sync cue occurs on the same frame number in every view.
\end{minipage}

Record a 1-minute test. Check for no dropped frames and that the synchronisation cue appears in every view (sample script available on GitHub - https://github.com/dimitrov-hristo/TrackStudio).

\subsubsection*{Usage}

\begin{figure}[!h]

\includegraphics[width=\textwidth]{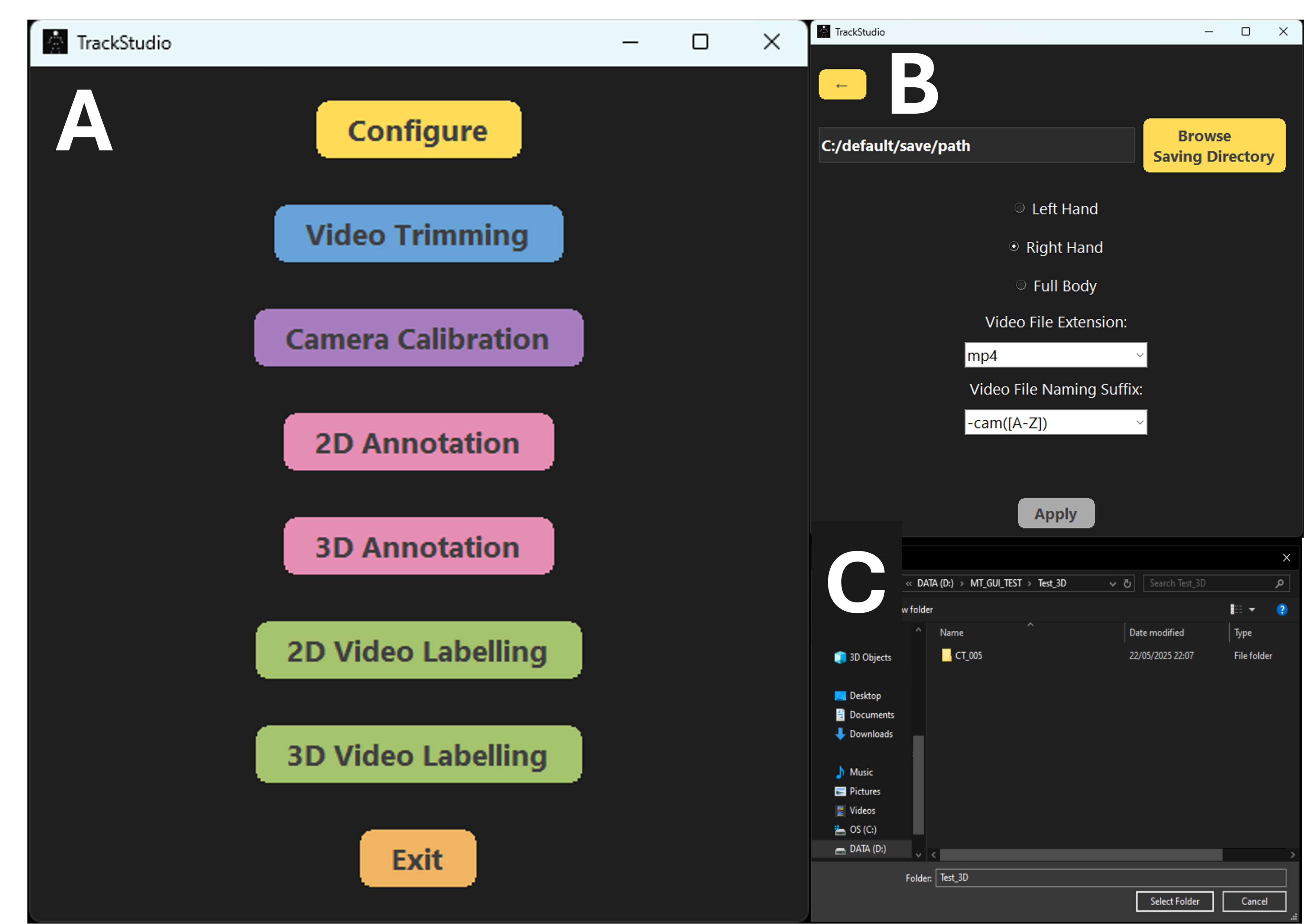}
\caption{TrackStudio GUI.
A: TrackStudio main window. B: Configuration section of TrackStudio GUI. C: Example of prompt to select input directory (e.g. for Automatic Video Trimming, 2D/3D Annotation, or 2D/3D Video Labelling).}
\label{suppfig4}

\end{figure}

TrackStudio (Supplementary Figure~\ref{suppfig4}) is operated through a graphical interface, but it does not enforce a step-by-step wizard. Hence, complete tasks in order according to your goal:
\begin{itemize}
    \item If you need 3D Annotation, you must first run 2D Annotation AND Camera Calibration. Make sure that camera views are time-aligned (see Synchronisation above).
    \item If you need to label/export 2D and/or 3D videos, you must first generate the 2D and/or 3D Annotations for them.
\end{itemize}

\subsubsection*{Selecting Saving Directory for TrackStudio}

\begin{figure}[!h]

\includegraphics[width=\textwidth]{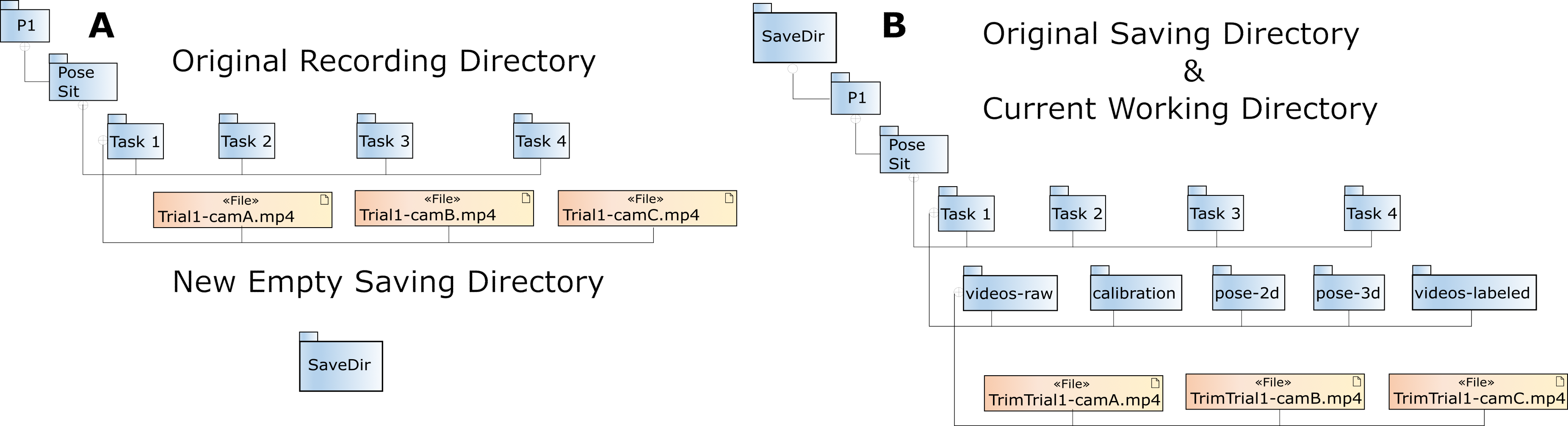}
\caption{Examples of the two cases for selecting a 'Saving Directory'.
A: An example of the beginning of the MLT process, where no steps have been executed and an empty folder is chosen as the 'Saving Directory'. B: An example of a 'Saving Directory' where some MLT work (calibration, 2D \& 3D annotation, and 2D video labelling) has already been done. If TrackStudio is re-opened and the user wants to complete, for example, 3D video labelling for all of the files that they have previously worked on, they should choose this 'Working Directory' as a 'Saving Directory'.}
\label{suppfig5}

\end{figure}

TrackStudio is designed to handle any folder structure you have for your data, as long as you have \textbf{video recordings only} in the final folder (e.g. 'Task 1', see Supplementary Figure~\ref{suppfig4}). When you choose a 'Saving Directory' in the GUI, there can be two instances:

\begin{enumerate}
    \item \textbf{First time using this folder – new Saving Directory} (Supplementary Figure~\ref{suppfig5}~A)
    \begin{itemize}
        \item If you have never run TrackStudio in the selected directory, the first processing step you run (e.g. Video Trimming or 2D Annotation) will:
        \begin{itemize}
            \item Recreate your original folder structure (e.g. participant → task → trial) inside the Saving Directory, and
            \item Saves trimmed videos there (if Video Trimming is first selected) or copies your video files into the directory (if 2D Annotation is selected).
        \end{itemize}
        \item Your original files stay where they are (Supplementary Figure~\ref{suppfig5}~A – 'Original Recording Directory'). TrackStudio works from the trimmed videos or copies (Supplementary Figure~\ref{suppfig5}~A – 'New Empty Saving Directory').
    \end{itemize}

    \item \textbf{Continuing in an existing MLT folder} (Supplementary Figure~\ref{suppfig5}~B)
    \begin{itemize}
        \item This folder already has results from Video Trimming and/or 2D Annotation and later steps.
        \item TrackStudio will use it as your working folder and save any new results alongside the existing ones.
        \item No new directories are created.
    \end{itemize}
\end{enumerate}

\vspace{0.5em}

\par\noindent\textit{Tips}:
\begin{itemize}
    \item Choose a fast drive (SSD) for your Saving Directory to reduce processing time.
    \item Make sure there is enough free space (videos are copied the first time) – TrackStudio will notify you if larger files are moved.
    \item Keep the lowest-level folders video-only to avoid import problems.
\end{itemize}

\subsubsection*{Steps for Markerless Tracking with TrackStudio}

TrackStudio includes several tools (“windows”). You only use the ones you need, but some steps depend on others:

\begin{itemize}
    \item 2D Video Labelling requires 2D Annotation first.
    \item 3D Annotation requires Camera Calibration and 2D Annotation.
    \item 3D Video Labelling comes after 3D Annotation.
    \item 2D/3D Video Labelling is optional - skip it if you do not need labelled videos for visualisation.
\end{itemize}

You can stop at any time and resume later; your progress is saved in your Saving Directory. For instance, after completing the Camera Calibration, you can close the GUI and proceed with 3D Annotation next time you open TrackStudio, by selecting the same Saving Directory.

When TrackStudio first opens, only Configure and Video Trimming are enabled. Complete the Configure step to unlock the other options.

Below are all the steps of MLT using TrackStudio:

\begin{enumerate}
    \item \textbf{(Always Necessary on Launch!) Configure} – every time you open the GUI, you need to complete a configuration step (see Supplementary Figure~\ref{suppfig4}~A and B), consisting of selecting:

    \begin{enumerate}
        \item \textbf{Saving Directory} - location where TrackStudio will store all your results (see 'Selecting Saving Directory for TrackStudio' above).
        \item \textbf{Body part to track} – Right Hand, Left Hand, or Full Body. If you want to track multiple body parts, then you will need to go through all the MLT steps for each body part separately.
        \item \textbf{Video file extension} – e.g. .mp4 or .mkv (match what you have recorded originally for your video files).
        \item \textbf{Camera name suffix} – the text at the end of the recording name that identifies each camera file. For example, if you have 3 cameras and your Trial1 video files are named 'Trial1–camA.mp4', 'Trial1–camB.mp4', and 'Trial1–camC.mp4', then the file naming suffix is going to be '–cam([A-Z])'.
    \end{enumerate}

    Once the configuration is completed click 'Apply' and go back to the Start Menu.

    \item \textbf{Video Trimming} – use this if you need to trim your videos and/or synchronise them using an LED(s) light. When you click Video Trimming, choose a mode:

    \begin{enumerate}
        \item \textbf{Manual}

        \begin{itemize}
            \item Pick a video, set start and end frames (use the arrow controls or type a number), then click \textbf{Trim}.
            \item The trimmed video is saved under videos-raw, inside your Saving Directory (TrackStudio mirrors your original folder structure).
            \item Repeat for the next video.
        \end{itemize}

        \item \textbf{Automatic (lights)} - see 'Best practices for LED synchronisation' above

        \begin{itemize}
            \item Based on detecting a red LED light turning ON at the start and OFF at the end of a trial/event.
            \item Choose a folder (a participant folder, a trial folder, or anything in between). TrackStudio will scan and trim all videos in subfolders.
            \item Click Select ROI (Region of Interest) – see Supplementary Figure~\ref{suppfig6}. For each camera view:

            \begin{itemize}
                \item A frame appears. Draw a small square around the LED.
                \item Press Enter to move to the next camera view and repeat.
                \item When all cameras are marked, you will see 'ROI Selected' – click \textbf{Trim}.
            \end{itemize}

            \item Trimmed videos are saved under "videos-raw" folder in your Saving Directory.

        \end{itemize}

        \underline{Advanced options for Automatic (lights) video trimming}

        \begin{itemize}
            \item Multiple trims within a video – set how many ON–OFF trials/events are inside one long recording. It will trim that number of videos.
            \item Recording length (in seconds) – enter a fixed duration if every trial/event has the same length (improves cut accuracy).
            \item Light On threshold (0–255) – set pixel light intensity threshold. Increase if your LED is dim; 255 is pure bright red.
            \item Pixel number threshold – set how many pixels need to go above the set light on threshold. Typically between 1 and 10 – higher means less false positives but also less sensitive to light changes.
        \end{itemize}
    \end{enumerate}

\begin{figure}[!h]

\includegraphics[width=\textwidth]{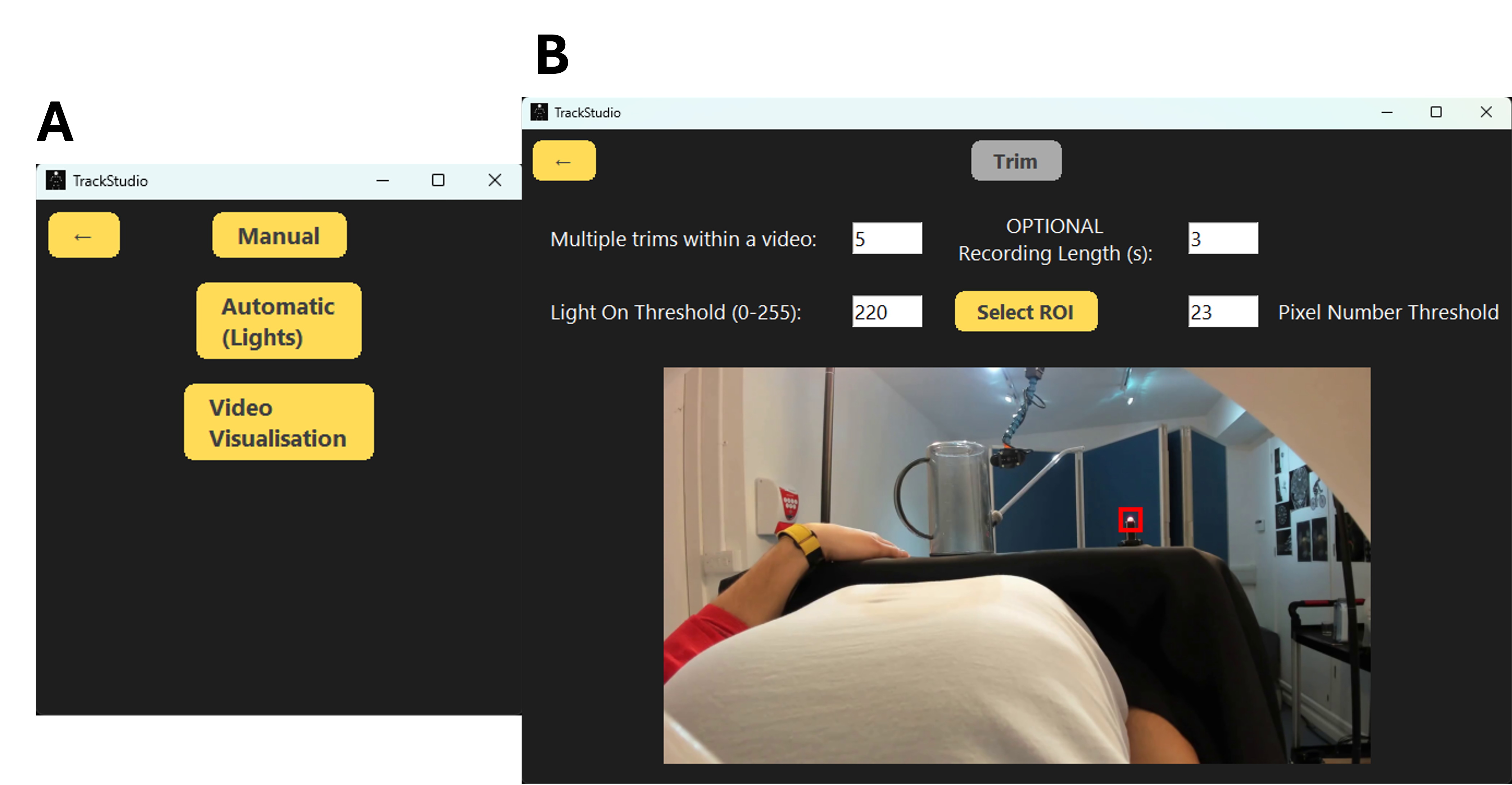}
\caption{TrakStudio’s Video Trimming. A: Video Trimming window, with Automatic (Lights) window, Manual, and Visualisation. B: Automatic trimming, illustrating LED region selection and optional trimming settings.}
\label{suppfig6}

\end{figure}

    \item \textbf{Camera Calibration (required for 3D)} – the Camera Calibration window, contains 2 sub-windows '\textbf{Change Board Parameters}' and '\textbf{Calibrate}':

    \begin{enumerate}
        \item \textbf{Change Board Parameters} – select only if you used a non-default calibration board (size, type, etc. – see Anipose documentation for details: https://anipose.readthedocs.io/en/latest/params.html). It will allow you to specify your calibration board parameters in a graphical way (no coding needed).

        \item \textbf{Calibrate} – select the folder with your calibration videos. TrackStudio will estimate each camera’s distortion and alignment parameters. During calibration:

        \begin{itemize}
            \item GUI window will display calibration progress messages.
            \item Once complete, final calibration error will be displayed
            \item A calibration file (calibration.toml) containing camera parameters is saved in a calibration/ folder inside your Saving Directory (and updated in TrackStudio). your saving directory as well as update this file in your TrackStudio directory. 
        \end{itemize}
    \end{enumerate}

    \textbf{\underline{IMPORTANT}}

    \begin{itemize}
        \item Do this before 3D Annotation.
        \item Quality matters: poor calibration footage can cause calibration to fail or result in highly inaccurate 3D Annotation. The Calibration Practical Recommendation section provides guidance on achieving reliable, high-quality calibration recordings.
        \item If you record multiple calibrations per participant (e.g. seated vs standing camera positions), you have to work in the sub-directory of each of the calibrations for 'Calibration' and the following '3D Annotation' and later steps for that participant. For example, run Standing calibration → 3D Annotation (standing tasks) before you do the Seated calibration. If you do Standing calibration → Seated calibration → 3D Annotation (standing tasks), TrackStudio will use the Seated calibration and the standing 3D results will be wrong.
    \end{itemize}

    \item \textbf{2D Annotation} – this step performs 2D MLT and creates files containing 2D coordinates of virtual markers corresponding to the body part you have chosen in the configuration section.

    \begin{itemize}
        \item Select a folder that contains your trimmed (and synchronised if 3D required later on) videos.
        \item Outputs are saved in your Saving Directory, mirroring your original structure and adding the needed subfolders for the underlying open-source tools. 

    \end{itemize}

    \item \textbf{3D Annotation} – combines 2D Annotation results across cameras (using your calibration) to produce 3D coordinates.

    \begin{itemize}
        \item Requirements: 2D Annotation and a matching Calibration completed for the selected folder directory.
        \item Select a folder (within your Saving Directory) that shares one calibration. For example:
        \begin{itemize}
            \item If Participant\_001 used one calibration for all tasks, select the Participant\_001 folder for calibration and 3D Annotation.
            \item If Participant\_001 used multiple calibrations within a session (e.g. seated vs standing), run Calibration and 3D Annotation per task within the task’s folder (e.g. \verb|Participant_001/Seated/|).
        \end{itemize}
    \end{itemize}

    \item \textbf{2D Video Labelling} (optional) – overlays 2D Annotation results onto their respective videos, visualising the results of the tracked body parts.

    \begin{itemize}
        \item Select the folder (within your Saving Directory) that contains the 2D Annotation results for the videos you want to review.
        \item Useful for quality checks; not required for 3D Annotation.
    \end{itemize}

    \item \textbf{3D Video Labelling} (optional) – creates a tiled video showing the original views with overlayed 3D Annotation results for visualisation.

    \begin{itemize}
        \item Select the folder (within your Saving Directory) that contains the 3D Annotation results for the videos you want to review.
        \item Note: if a recording used more than 4 cameras, the combined preview shows overlays on the first 4 views only (a limitation of the underlying video tiling library). Your 3D data is still computed from all the cameras you used.
    \end{itemize}
\end{enumerate}

For advanced users and developers, the GitHub page explains how each script works and how to adjust or extend TrackStudio (e.g. to swap in another 2D/3D tracking method). 

\subsubsection*{Practical Recommendations}

\underline{Visibility \& Camera Views}

\begin{itemize}
    \item Minimise occlusions whenever possible.
    \item Do a short test, recording complete task movements, and decide how many cameras are needed to fully capture all movements in the space.
    \item For 3D MLT, the tracked body part must be visible in at least two camera views at all times. Hence, at least 2 cameras are needed but it is recommended to use 3 or more.
    \item Full-body or complex movements will require a higher number of cameras and a larger space for recording, with cameras evenly distributed around the room and positioned at different heights.
\end{itemize}

\vspace{0.5em}

\par\noindent\underline{Frame Rate \& Hardware}

\begin{itemize}
    \item For rapid movements, high frame rate cameras (120–240 fps) are advisable to reduce motion blur and maintain stable tracking.
    \item More/higher-end cameras increase load: you will need fast USB 3.x ports, a strong CPU, and often a GPU for smooth preview (see 'Video Recordings').
    \item If the system cannot keep up, you will get dropped frames (lost data).
\end{itemize}

\vspace{0.5em}

\par\noindent\underline{Lighting \& Clothing}

\begin{itemize}
    \item Keep the space well-lit and avoid shadows, especially during calibration.
    \item Avoid loose clothing or patterned fabrics – they may interfere with body landmark detection, similar to challenges in optical tracking.
    \item Prefer simple, well-fitting clothes (e.g. scrubs), especially for full-body recordings
\end{itemize}

\vspace{0.5em}

\par\noindent\underline{Recommendations for Improving Calibration Video Recordings:}

\begin{enumerate}
    \item Ensure the environment is well-lit. Poor lighting can prevent the calibration board from being properly detected.
    \item Avoid reflections on the printed calibration board. Glossy paper, clear tape, or laminating the board may introduce reflections that interfere with detection.
    \item Verify the calibration board is printed at the correct scale and with high contrast. Some printers automatically resize prints to fit the page, which may compromise calibration accuracy. The dark areas of the board should be fully black to enhance feature detection.
    \item Leave sufficient margins for holding the board. This prevents hand occlusions. Mounting a handle at the back of the board is recommended.
    \item Use a ChArUco calibration board. This type is more tolerant to occlusions and partial views. Not all board corners must be visible in every camera view.
    \item We recommend recording calibration videos for at least 1 minute per camera. For example, with three cameras, a total calibration duration of approximately 3 minutes is advised.
    \item Monitor the recording during calibration. Position the board so that all cameras have an unobstructed view. This should be your starting and ending position.

    \begin{enumerate}
        \item Begin recording and keep the board visible to all cameras for approximately 20 seconds.
        \item Then, move the board closer to one camera at a time. Each up-close view should be visible to only one camera at a time (avoid partial board views in neighbouring cameras).
        \item During the up-close phase, hold the board steady, then gently rotate it to capture different angles and lighting conditions. Spend about 40 seconds to 1 minute per camera.
        \item After each close-up, slowly move the board backwards until it is fully visible in two cameras. Hold it in this position briefly and rotate it slightly to capture various perspectives (while avoiding partial detections).
        \item Repeat the process for each camera so that all cameras record both up-close and shared views (e.g., A+B, B+C, A+C).
        \item Finally, return to the starting position where all cameras can see the full board, and slowly rotate it again.
    \end{enumerate}

\end{enumerate}

\end{document}